\documentclass[11pt,a4paper]{article}
\usepackage[hyperref]{emnlp2020}
\usepackage{times}
\usepackage{latexsym}

\usepackage{microtype}

\usepackage{amsmath}
\usepackage{amsfonts,amssymb}
\usepackage{bbm}
\usepackage{booktabs}
\usepackage{graphicx}

\usepackage{array}
\newcolumntype{L}[1]{>{\raggedright\let\newline\\\arraybackslash\hspace{0pt}}m{#1}}
\newcolumntype{C}[1]{>{\centering\let\newline\\\arraybackslash\hspace{0pt}}m{#1}}
\newcolumntype{R}[1]{>{\raggedleft\let\newline\\\arraybackslash\hspace{0pt}}m{#1}}
\usepackage{multirow,tabularx}

\usepackage{xcolor}

\newcommand{\blueuline}[1]{{\color{cyan}\underline{{\color{black}#1}}}}
\newcommand{\red}[1]{{\color{red}#1}}
\definecolor{almond}{rgb}{0.99, 0.96, 0.89} 
\aclfinalcopy 
\def\aclpaperid{3184} 

\title{Partially-Aligned Data-to-Text Generation with Distant Supervision}

\author{Zihao Fu\textsuperscript{\rm 1}, Bei Shi\textsuperscript{\rm 2}, Wai Lam\textsuperscript{\rm 1}, Lidong Bing\textsuperscript{\rm 3}, Zhiyuan Liu\textsuperscript{\rm 4} \\
\textsuperscript{\rm 1}Department of Systems Engineering and Engineering Management, \\ The Chinese University of Hong Kong \\
\textsuperscript{\rm 2}Tencent AI Lab 
\textsuperscript{\rm 3}DAMO Academy, Alibaba Group  \\
\textsuperscript{\rm 4}State Key Lab on Intelligent Technology and Systems\\
Department of Computer Science and Technology\\
Institute for Artificial Intelligence, Tsinghua Univerisity\\
zhfu@se.cuhk.edu.hk; beishi@tencent.com; wlam@se.cuhk.edu.hk; \\l.bing@alibaba-inc.com; liuzy@tsinghua.edu.cn
}

\date{}

\begin{document}

\maketitle
\begin{abstract}
The Data-to-Text task aims to generate human-readable text for describing some given structured data enabling more interpretability.
However, the typical generation task is confined to a few particular domains since it requires well-aligned data which is difficult and expensive to obtain. Using partially-aligned data is an alternative way of solving the dataset scarcity problem. This kind of data is much easier to obtain since it can be produced automatically. However, using this kind of data induces the over-generation problem posing difficulties for existing models, which tends to add unrelated excerpts during the generation procedure.
In order to effectively utilize automatically annotated partially-aligned datasets, we extend the traditional generation task to a refined task called Partially-Aligned Data-to-Text Generation (PADTG) which is more practical since it utilizes automatically annotated data for training and thus considerably expands the application domains.
To tackle this new task, we propose a novel distant supervision generation framework. It firstly estimates the input data's supportiveness for each target word with an estimator and then applies a supportiveness adaptor and a rebalanced beam search to harness the over-generation problem in the training and generation phases respectively. We also contribute a partially-aligned dataset \footnote{The data and source code of this paper can be obtained from \url{https://github.com/fuzihaofzh/distant_supervision_nlg}} by sampling sentences from Wikipedia and automatically extracting corresponding KB triples for each sentence from Wikidata.
The experimental results show that our framework outperforms all baseline models as well as verify the feasibility of utilizing partially-aligned data.
\end{abstract}

\section{Introduction}
The Data-to-Text generation task focuses on generating human-readable text corresponding to some given structured data. For example, given the input knowledge base (KB) triple $\langle$Company of Heroes, developer, Relic Entertainment$\rangle$, the aim is to generate a text description such as ``\textit{Company of Heroes is developed by Relic Entertainment.}''. In recent years, many works have been proposed to give impetus to the Data-to-Text generation task. For instance,  \citeauthor{gardent2017creating} \shortcite{gardent2017creating,gardent2017webnlg} proposed the WebNLG task aiming at generating description text of the given KB triples. \citet{novikova2017e2e} proposed the E2E task aiming at generating restaurant reviews according to the given restaurant attributes. \citet{lebret2016neural} proposed the WikiBio task in which the biography of each person is generated according to the given Wikipedia infobox. 

\begin{figure}[t]
\centering
\includegraphics[width=1\columnwidth]{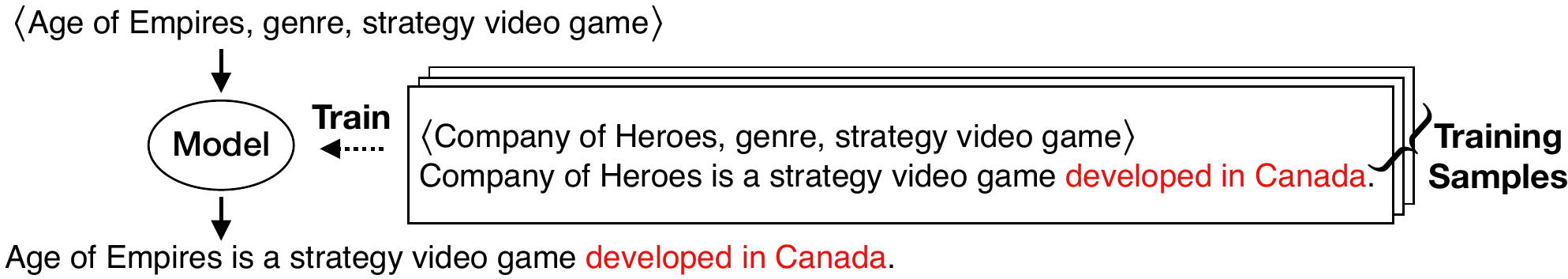}
\caption{Illustration of the over-generation problem in the partially-aligned data-to-text generation task. In the training set, there is no KB triple corresponding to the text ``\textit{developed in Canada}''. The model is likely to bind the text to existing triples incorrectly. As a result, during the testing or operational stage, the model is likely to overly generate this kind of excerpt for similar triples.}
\label{fig:superfluous} 
\end{figure}

These typical data-to-text generation tasks are confined to a few particular domains since it requires well-aligned data and text pairs which are difficult and costly to obtain. Specifically, it is required that each input data provides exactly the same information with the target text. This requirement makes the dataset difficult to build and confines the task to particular domains where such kind of data (WikiBio, E2E) or human-labeled data (WebNLG) are available. 
Using partially-aligned data is an alternative way of solving the dataset scarcity problem. Partially-aligned data do not require that each part in the text is exactly aligned with a particular input KB triple. This kind of data is much easier to obtain with automatic methods. Consequently, it can handle much broader kinds of domains. However, it induces the over-generation problem\footnote{Note that omission problem also happens in the Data-to-Text generation task, but it is not a major problem.}. As shown in Fig.~\ref{fig:superfluous}, some parts (``\textit{developed in Canada}'') in the generated text for $\langle$Age of Empires, genre, strategy video game$\rangle$ have not been mentioned in the input KB triple. Essentially, it is because in the training set, such unrelated text exists in some training samples. During the training, it misleads the model to bind the text ``\textit{developed in Canada}'' to some irrelevant KB triples. When similar triples exist in the testing, it is prone to adding some over-generated text which is actually unrelated to the given input data. Current generation models fail to be trained on such partially-aligned data due to lacking the tolerance of the over-generation problem. 

In order to effectively utilize automatically annotated partially-aligned datasets for handling more domains, we extend the traditional generation task to a refined task called Partially-Aligned Data-to-Text Generation (PADTG). Like the traditional task, the PADTG task also requires generating text with respect to the given input data. However, for the training data, we only require that the given structured data contains partial information of the corresponding text. This task is more practical since it utilizes the partially-aligned data for training and thus considerably expands the application domains. However, due to such data's nature, successfully suppressing the over-generation problem is the critical point for proposing an effective model. 

We propose a Distant Supervision Generation (DSG) framework to tackle the PADTG task. Our framework can deal with the challenging over-generation problem when training on the partially-aligned data. It firstly trains an estimator to calculate each word's supportiveness in the target sentence with respect to the input data, i.e. how likely the word is conveyed by the input triples. Then the framework employs a sequence-to-sequence (S2S) neural model to encode the input data and generates the description sentence accordingly. In the training procedure, a supportiveness adaptor is used to adapt the estimated supportiveness into the loss function while in the generation procedure, a rebalanced beam search is used to generate text augmented with the supportiveness scores.

To prepare the partially-aligned data, we build a new dataset called WITA from text sources, namely, Wikipedia and Wikidata. We propose a novel KB extractor to extract KB triples given a piece of text sampled from Wikipedia. The KB extractor firstly detects named entities with an entity detector. The triple retriever queries the Wikidata database to find the most matching triples corresponding to these entities. We filter the results with a matching score to remove unextractable sentences. 

Our contributions can be summarized as follows. (1) We propose a new task, namely, partially-aligned Data-to-Text generation, which is more practical and extensible to more domains. (2) We propose a distant supervision generation framework that can tackle the challenges of the new task including the over-generation problem. (3) We contribute a sizeable partially-aligned dataset suitable for this task.

\section{Method}

\subsection{Overview}
Formally, we denote the input KB triples as $K=[\langle h_1,r_1,t_1\rangle, \cdots, \langle h_n,r_n,t_n\rangle]$, where $h_i,r_i,t_i$ represent the $i$th head, relation, and tail respectively while $n$ is the number of triples. The corresponding text is denoted as $T=[w_1,\cdots,w_m]$, in which $w_i$ is the $i$th word in $T$ and $m$ is the sentence length. It should be noted that, in the task of Partially-Aligned Data-to-Text  Generation (PADTG), $T$ contains some information that $K$ does not have.
The target of the task is to train a model that generates text $T'$ that exactly describes the KB triples in $K$. 

\begin{figure}[t]
\centering
\includegraphics[width=0.9\columnwidth]{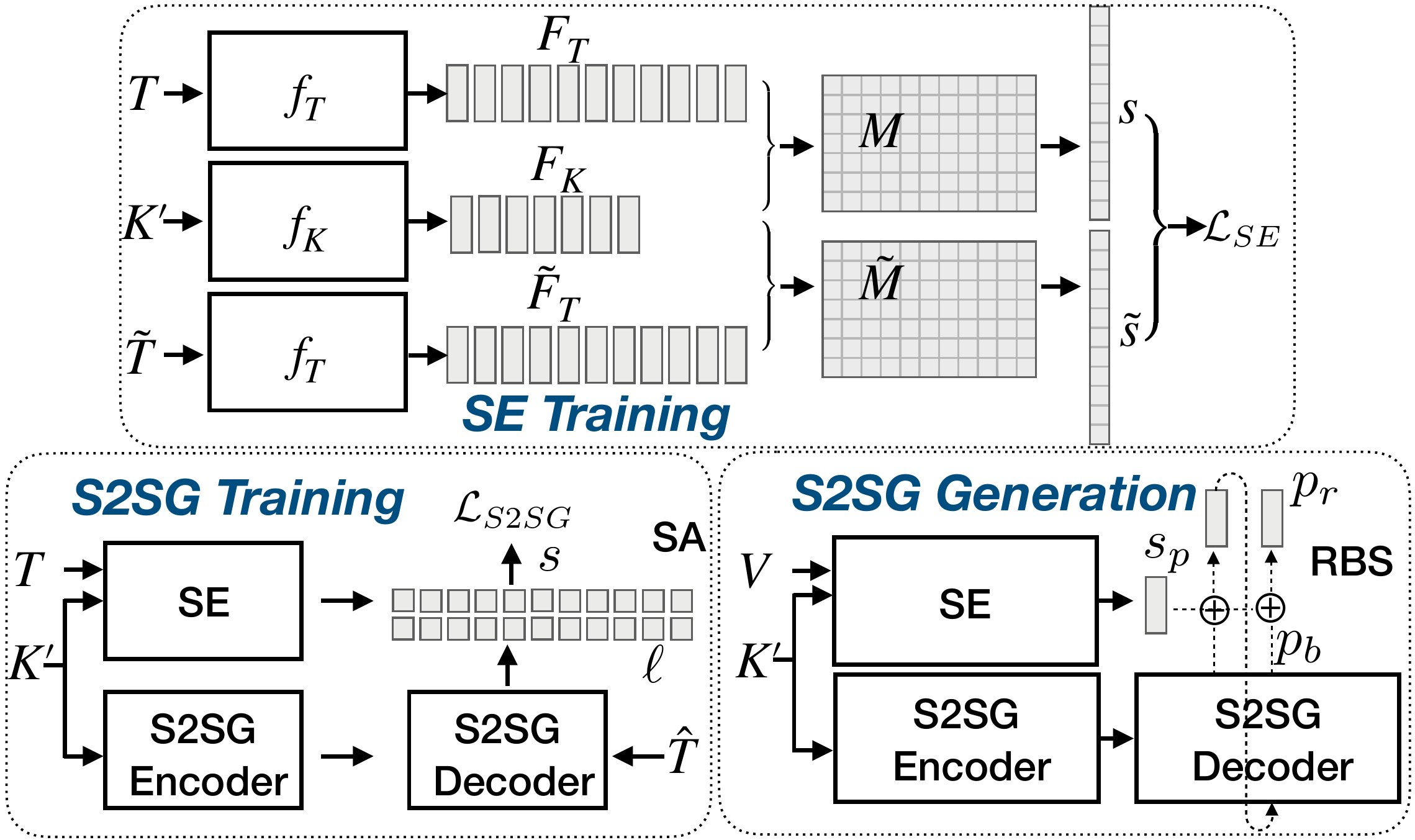}
\caption{Distant Supervision Generation Framework.}
\label{fig:framework} 
\end{figure}

Our proposed Distant Supervision Generation (DSG) framework contains four components, namely a Supportiveness Estimator (SE), a Sequence-to-Sequence Generator (S2SG), a Supportiveness Adaptor (SA), and a Rebalanced Beam Search (RBS). 
As illustrated in Fig.~\ref{fig:framework}, in the SE training procedure, we first pre-train the SE component to estimate a supportiveness vector $s\in \mathbb{R}^m$ indicating whether each target word $w_i\in T$ is describing the input triples in $K$. It adopts the self-supervised mechanism that trains the model to maximize the margin between the target words' scores and negative sampled words' scores. Then, the pre-trained SE component is utilized to estimate a supportiveness vector $s$ in both S2SG Training and S2SG Generation. 
In the S2SG training procedure, the S2SG model firstly calculates the generation loss $\ell$. Then, SA combines $\ell$ with $s$ to get a refined loss in which the loss is diminished if one target word has lower supportiveness. 
In the S2SG generation procedure, the RBS component combines $s$ with the probability distribution of candidate words to obtain a better generation result.

\subsection{Supportiveness Estimator}
We concatenate the input KB triples word-by-word as $K'=[w_1^{h_1},\cdots,w_{|h_1|}^{h_1},$ $w_1^{r_1},\cdots,w_{|r_1|}^{r_1},$ $w_{1}^{t_1},\cdots,w_{|t_1|}^{t_1}, $ $\text{KBSEP}, w_1^{h_2}$ $\cdots \cdots,w_{|t_n|}^{t_n}]$, in which $h_j,r_j,t_j$ is the head, relation and tail entity of the $j$th triple. $w_{i}^{h_j}$ represents the $i$th word in the $j$th KB triple's head entity. $|h_j|$ stands for the word count of the $j$th head entity. KBSEP is a seperator between each triple.

 \textbf{Feature Extraction}. In SE, a feature extraction component $f_K$ is utilized to extract features for each word denoted as $F_K=f_K(K')$, in which $F_K\in \mathbb{R}^{d \times |K'|}$ is the extracted feature matrix for $K'$. $d$ is the embedding dimension and $|K'|$ is the length of $K'$. Specifically, in the feature extraction component, $K'$ is firstly embedded with an embedding layer as $K_1 = \text{emb}(K')$, $K_1\in \mathbb{R}^{d \times |K'|}$ . Then $K_1$ is sent into a normalization layer \cite{ba2016layer} as $K_2=\text{NL}(K_1)$, $K_2\in \mathbb{R}^{d \times |K'|}$. NL is defined as $$\text{NL}(x)=\frac{\gamma(x-\mathbb{E}(x))}{\sqrt{\text{Var}(x)+\epsilon}}  +\beta,$$ in which $\mathbb{E}$ and $\text{Var}$ are mean and variance of the input $x$ while $\gamma$ and $\beta$ are learnable parameters. $\epsilon$ is a small constant which is usually set to $1.0e-5$. $K_2$ is then sent into a combination of linear feedforward layers and a ReLU layer as $K_3=\text{FW}_2(\text{ReLU}(\text{FW}_1(K_2)))$, $K_3\in \mathbb{R}^{d \times |K'|}$, where $\text{FW}_2$ and $\text{FW}_1$ are linear feedforward layer while $\text{ReLU}$ stands for the ReLU layer. Afterwards, the features representation is calculated as $F_K=\text{NL}(K_3)$. Similarly, the features for each word in the target text is denoted as $F_T=f_T(T)$, $F_T\in \mathbb{R}^{d \times m}$. 

\textbf{Supportiveness Vector}. We calculate the supporting matrix as $M=F_K^T F_T$, $M\in\mathbb{R}^{|K'|\times m}$, in which $M_{i,j}$ represents the supportiveness of the $i$th word in $K'$ that support for the $j$th word in $T$. The supportiveness score vector is aggregated from $M$ as $$s_j=\log\sum_{i=1}^{|K'|}\exp(M_{i,j}),$$ where $s_j$ is the $j$th element of the vector $s\in\mathbb{R}^{m}$ and it stands for input $K$'s supportiveness to the $j$th word. 

\textbf{Negative Sampling}. In order to prevent the model from giving all words a high supportiveness score, we use the negative sampling method to sample some negative sentences. We denote the empirical distribution of the words in the target text as $P_{\mathcal{T}}$ in which $\mathcal{T}$ is the set of all target sentences. We sample words from $P_{\mathcal{T}}$ while avoiding sampling the same words in T. The sampling procedure can be denoted as $\tilde{w}_i\sim P_{\mathcal{T}},\tilde{w}_i \not \in T$. The negative sample is composed of $\tilde{w}_i$s as $\tilde{T}=[\tilde{w}_1, \cdots,  \tilde{w}_{m}]$, where $\tilde{w}_i$ is the $i$th word in $\tilde{T}$ which has the same length as $T$. The negative sample $\tilde{T}$ will also be fed to the network in the same way as the original target $T$. The supportiveness score vector for the negative sample is denoted as $\tilde{s}\in\mathbb{R}^{m}$.

\textbf{Optimization Target}. The overall loss function consists of a margin loss, a word-consistent loss, and a concentration loss. The \textbf{margin loss} is defined as the margin between the supportiveness of the original text and that of the negative sample, which can be written as $$\mathcal{L}_m=\sum_{i=1}^{\tilde{m}}\sigma(\tilde{s}_i)-\sum_{i=1}^{m}\sigma(s_i),$$ in which $\sigma(x)=1/(1+e^{-x})$ is a sigmoid function. Minimizing $\mathcal{L}_m$ helps maximize the gap between the positive and the negative samples. The \textbf{word-consistent loss} is used to make the supportiveness from the same word in input KB larger than the supportiveness from different words. It is defined as $$\mathcal{L}_w=-\sum_{i=1}^{m}\sum_{j=1}^{|K'|}\mathbbm{1}(T_i=K'_j)[M_{i,j}-\log(\sum_{k=1}^{|K'|}\exp M_{k,j})].$$ It increases the supportiveness $M_{ij}$ if the $i$th word in $T$ and the $j$th word in $K'$ are the same word.  
The \textbf{concentration loss} is used to avoid one word in $K'$ supporting too many words in $T$. It is denoted as $$\mathcal{L}_c=\max_i \sum_{j=1}^{m}M_{i,j}.$$ If one word supports too many words, all its corresponding supportiveness will be penalized. The overall loss function is denoted as the weighted sum of these loss functions as $$\mathcal{L}_{SE}=\mathcal{L}_m+\omega_w \mathcal{L}_w+\omega_c \mathcal{L}_c,$$ in which $\omega_w$ and $\omega_c$ are tunable hyper-parameters.

\subsection{Sequence-to-Sequence Generator}
We use the Transformer \cite{vaswani2017attention} structure as our S2S generator. The Transformer is an attention-based structure and is widely used in many tasks. It contains two major components, namely an encoder and a decoder. All of these components are built with several attention layers. The encoder firstly takes $K'$ as input to generate the feature representation: $G_K=\text{Enc}(K')$. The decoder takes shifted target text $\hat{T}$ (shift last EOS tag to the beginning) as input and get the negative log-likelihood for each word in $T$ as $\ell=\text{Dec}(\hat{T},G_K)$, $\ell\in \mathbb{R}^{|T|}$, where $\text{Dec}$ is the transformer decoder and $|T|$ is the length of $T$.
We refer readers to \citet{vaswani2017attention} for more technical details.

\subsection{Supportiveness Adaptor}
The supportiveness Adaptor adapts the supportiveness score to S2SG's output. We investigate three methods, namely, Hard Adaptor, Soft Adaptor, and Attention Adaptor.

\textbf{Hard Adaptor}. With the supportiveness scores, we can simply remove words that have low supportiveness. For each word $w_i$ in the target sentence $T$, we use a uniform random number generator to generate a random number $r_i\in [0, 1]$. we ignore words in $T$ if $r_i > s_i$ and copy it into $T'$ otherwise. Then, $T'$ is used instead of $T$ in the training procedure.

\textbf{Soft Adaptor}. Since the hard adaptor directly removes words, it is easy to omit essential words and make the model generating unreadable text. We propose a soft adaptor to alleviate this issue. We use the original target text $T$ as input. For the output negative log-likelihood loss vector $\ell$, we combine it with the supportiveness vector $s$ to modify the S2SG's loss as $$\mathcal{L}_{S2SG}=\sum_{i=1}^m \ell_i s_i.$$

\textbf{Attention Adaptor}. Instead of using SE to estimate the supportiveness vector, attention adaptor directly aggregates the attention matrix as the supportiveness scores in our proposed DSG model. For each target word, it takes its max attention weight on each source word as the supportiveness score. We use maximization to aggregate the scores instead of considering all scores because all attention weights sum up to 1, and thus irrelevant words can also be assigned some attention. Using maximization aggregation avoids such irrelevant words. The supportiveness scores are then utilized in a similar way as the soft adaptor. 

\subsection{Rebalanced Beam Search}
In the generation step, the supportiveness scores can also be utilized to help rebalance the final word probability distribution. We make a pseudo target sequence as $T_p=[1,2,\cdots,|V|]$ which contains all words in vocabulary $V$. The supportiveness score $s_p\in\mathbb{R}^{|V|}$ for all words is calculated as the same procedure in training. In the traditional beam search, it outputs a probability $p_b\in\mathbb{R}^{|V|}$ over the whole vocabulary denoting the possibility for each token in the vocabulary to be chosen as the next word. We rebalance the probability with the supportiveness score vector as $p_r=p_b\cdot s_p^\alpha$ where $\alpha$ is a tunable hyper-parameter.

\section{WITA: Our Partially-Aligned Dataset}
\begin{figure}[t]
\centering
\includegraphics[width=0.99\columnwidth]{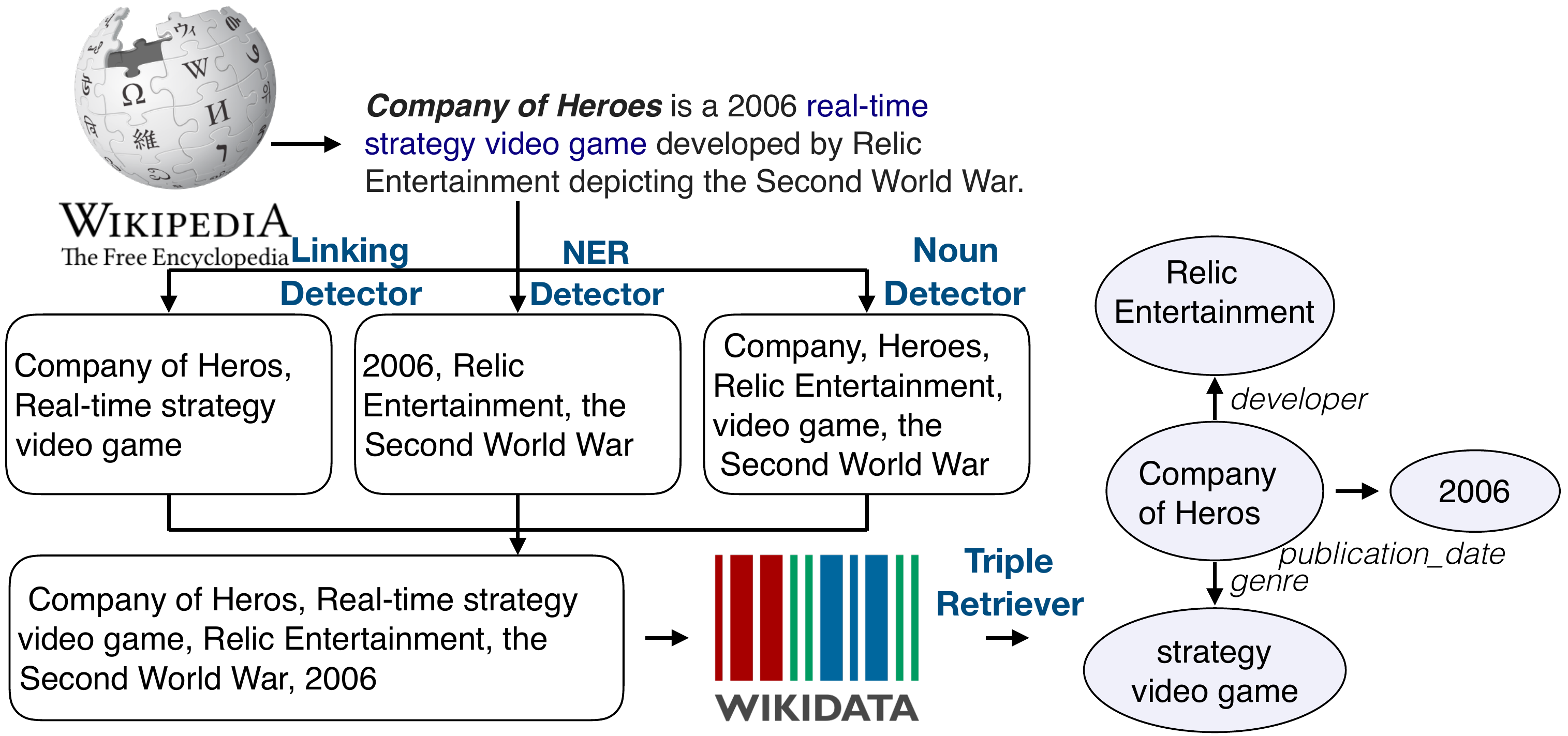}
\caption{Our proposed KB extractor for harvesting the partially-aligned data from Wikipedia and Wikidata.}
\label{fig:make-dataset} 
\end{figure}

We automatically harvest some partially-aligned data from Wikipedia and Wikidata and prepare a dataset called WITA.
We select each Wikipedia article's first sentence from the 20190920 Wikipedia dump\footnote{https://dumps.wikimedia.org/enwiki/} as the target text. Then, we remove irrelevant tags and links with several predefined rules.
We propose a KB extractor, as illustrated in Fig.~\ref{fig:make-dataset}, which can take the selected Wikipedia sentences and extract the corresponding KB triples. In the KB extractor, named entities are detected by an entity detector. The detected named entities are then combined into pairs by the Cartesian Product operation. The triples that mention these entity pairs are retrieved by a triple retriever that searches the corresponding KB triples from the Wikidata database. We use an entity-recall based score to filter inappropriate sentences.

\subsection{Entity Detector}
We use three sub-detectors to recognize named entities and union them together. We first use a NER detector based on the spaCy\footnote{https://spacy.io/}'s NER tool to recognize the named entities. Then we use the Noun detector based on spaCy's noun chunks recognition component to identify noun chunks. This detector is used because noun chunks have a high probability of being named entities. Finally, we use a linking detector, which is rule-based, to extract entities tagged with internal links. The detected entities for given sentence $c$ is denoted as $E_c=[e_1,e_2,\cdots,e_p]$, while $p$ is the entity number.

\subsection{Triple Retriever}
In order to quickly retrieve related triples for given named entities, we first store the Wikidata database in Elasticsearch\footnote{https://www.elastic.co/products/elasticsearch}. We concatenate all possible variant names for an entity in Wikidata as the entity name. For example, \textit{Steve Jobs}'' has alternative names like ``\textit{Steven Paul Jobs}'' and ``\textit{Steven Jobs}''. The entity name is concatenated as ``\textit{Steve Jobs | Steven Paul Jobs | Steven Jobs}''. In the Wikidata database, each triple contains a head, a relation and a tail entity and we denote the set of all the preprocessed triple as $\mathcal{D}=\{\langle h_i,r_i,t_i\rangle|\forall i \}$, where $h_i,r_i,t_i$ are the head, relation and tail entity for the $i$th triple.

For given detected entities $E_c$, we make a list of named entity pairs by conducting a Cartesian Product as $C_e=\{\langle e_i,e_j\rangle | \forall e_i \in E_c, e_j\in E_c,e_i\ne e_j\}$. Afterwards, we query the Wikidata database to find a triple that matches the given named entity pair $\langle e_i, e_j\rangle\in C_e$ to make the head entity close to $e_i$ while the tail entity close to $e_j$. It should be noted that the relation may be wrong, i.e. the matching triple describes a relation different from the one in the input sentence. But in reality, this probability is very small since most of the entity pairs only have one corresponding relation. For a given named entity pair $\langle e_i, e_j\rangle$, the query condition can be formally expressed as:
$$\begin{aligned}
    &\widehat{\langle h,r,t\rangle}=\arg\max_{\langle h,r,t\rangle} d (g(e_i,h) + g(e_j,t)) + \\&\ \ \ \ \ \ \ \ \ \ \ \ \ \ \ \ \  (1-d)(g(e_j,h) + g(e_i,t))\\
&\text{s.t.} \ \ (1-d)M + l(e_i,h)\geqslant\kappa;\ \ dM + l(e_i,t)\geqslant\kappa; \\
&\ \ \ \ \ \ \ (1-d)M + l(e_j,t)\geqslant\kappa;\ \ dM + l(e_j,h)\geqslant\kappa;\\
&\ \ \ \ \ \ \ \langle h,r,t\rangle\in \mathcal{D};\ \ \ d\in \{0,1\},
\end{aligned}$$
in which $g$ is a single-term matching score\footnote{https://www.elastic.co/guide/en/elasticsearch/guide/current/ practical-scoring-function.html} while $l$ is the string similarity metric ranging from 0 to 1. $M$ is a sufficiantly large number and $d$ is an integer. $\kappa$ is a threshold preventing the retrieved head and tail being too different from $e_1$ and $e_2$.
After we have retrieved entities for all sentences, we calculate a score based on entity-recall to filter wrongly extracted data-text pairs. entity-recall for KB triples and the corresponding text is defined as 
$$\begin{aligned} r_e=\sum_{j=1}^m\mathbbm{1}\{\sum_{i=1}^n [&\mathbbm{1}(w_j\in h_i) + \mathbbm{1}(w_j\in r_i) + \\ & \mathbbm{1}(w_j\in t_i)]>0\}/m,
\end{aligned}$$ where $m$ is the length of the sentence while $n$ is the triple number. $r_e$ indicates how much information in text has been covered by the retrieved triples.

Since WebNLG is the most similar task to our PADTG task among others, we compare the statistics of our WITA dataset with WebNLG in Table~\ref{tab:wita}. It can be observed that (1) WITA is larger than the WebNLG dataset making it more practical. It can be easily extended to more domains. (2) WITA contains more relation types and entity types than that of WebNLG, indicating that our dataset involves more domains. (3) The vocabulary of the target sentences of WITA is much larger than that of WebNLG, which shows that our dataset is more challenging and more realistic. (4) The entity-recall score of WITA is lower than WebNLG. This is because WITA is automatically annotated and some information in the text is not contained in the KB triples. The low entity-recall score causes the over-generation problem and the specific value measures how serious the problem is.

\begin{table}[t]
  \centering
  \small
  \begin{tabular}{@{~}l@{~}@{~}l@{~}@{~}l@{~}}
\toprule
{} &                     WITA &                  WebNLG \\
\midrule
Size             &                    55,400 &                   42,892 \\
Relation Type    &                      640 &                     373 \\
Entity Type       &                   128,405 &                    3,114 \\
Text Length         &     (18.8, 17.0, 5, 59) &   (23.7, 22.0, 4, 84) \\
KB Number           &  (3.0, 3.0, 1, 11) &  (2.9, 3.0, 1, 7) \\
Vocabulary            &                   102,404 &                    8,886 \\
entity-recall    &                    0.508 &                   0.625 \\
\bottomrule
\end{tabular}
  \caption{Statistics of WITA and WebNLG. For the text length and KB number, the data are mean, median, min and max respectively.}
  \label{tab:wita}
\end{table}

\section{Experiments}
\subsection{Experimental Setup}
We split WITA into a training set, a development set, and a testing set of 50,000, 5,000, and 400 records respectively. For the purpose of evaluating the performance of the models, we ask human helpers to annotate the testing set sentences. The human helpers are asked to revise the input KB triples and the corresponding target sentences making them exactly consistent with each other.
We use several evaluation metrics including BLEU \cite{papineni2002bleu}, ROUGE$_L$ \cite{lin2004rouge}, METEOR \cite{banerjee2005meteor}, NIST \cite{doddington2002automatic} and CIDEr \cite{vedantam2015cider} with the package provided by \citet{novikova2017e2e}. We follow the default setting in ROUGE$_L$ where $\beta$ is set to 1.2. We build our model based on the Transformer model \cite{vaswani2017attention,ott2019fairseq} and use Byte Pair Encoding (BPE)  \cite{sennrich2016neural} to build the subword dictionary. We use Fairseq \cite{ott2019fairseq} to build our model and keep all hyper-parameters for Transformer unchanged. We set $\kappa=0.75$ from $\{0.1,0.25,0.5,0.75,0.9\}$ by extracting samples and ask human helper to evaluate. We use grid search to tune hyper-parameters on the development set and choose $\omega_w=0.05$ from \{0.02,0.05,0.1,0.2,0.5,1.0,2.0,5.0\}, choose $\omega_c=1.0$ from \{0.02,0.05,0.1,0.2,0.5,1.0, 2.0,5.0\} and choose $\alpha=0.1$ from \{0.02,0.05,0.1,0.2,0.5,1.0\}. The model has 49M parameters and it takes 2.4 hours to train it on a NVIDIA TITAN RTX graphics card.

\subsection{Comparison Models}
We compare our full DSG model with the following baselines, state-of-the-art models, and ablations.

\textbf{S2S} utilizes the traditional S2S model \cite{sutskever2014sequence,cho2014learning} equipped with attention \cite{bahdanau2014neural,luong2015effective} and copy \cite{gu2016incorporating} mechanism. It is recognised as the state-of-the-art model \cite{shimorina2018handling} in the WebNLG \cite{gardent2017webnlg} task. 

\textbf{S2ST} utilizes the prevalent Transformer model \cite{vaswani2017attention,ott2019fairseq} which outperforms the traditional S2S model in many generation tasks. 

\textbf{DSG-A} utilizes the attention adaptor which adapts attention as the supportiveness scores in the loss.

\textbf{DSG-H} is almost the same as our DSG model. The only difference is that the supportiveness scores are adapted with hard adaptor while our DSG model uses the soft adaptor.

\textbf{DSG w/o RBS} is an ablation model. It removes the Rebalanced Beam Search component from our DSG model.

\textbf{DSG w/o SA} is an ablation model without the Supportiveness Adaptor. The supportiveness scores are only used by RBS in the generation phase. They are not adapted to the training loss.

\subsection{Experimental Results}

\begin{table}[t]
  \centering
  \small
\begin{tabular}{@{~}l@{~}@{~}l@{~}@{~}l@{~}@{~}l@{~}@{~}l@{~}@{~}l@{~}}
\toprule
{} &   BLEU &  NIST & METEOR & ROUGE$_L$ & CIDEr \\
\midrule
S2S                  &  0.463 &  7.97 &  0.385 &   0.693 &  4.12 \\
S2ST              &  0.496 &  8.05 &  0.417 &   0.721 &  4.53 \\
DSG-A     &  0.518 &  8.36 &  0.421 &    0.730 &  4.75 \\
DSG-H    &    0.500 &  8.61 &  0.403 &   0.711 &  4.65 \\
DSG           &  \textbf{0.555} &  \textbf{8.71} &   \textbf{0.425} &   \textbf{0.742} &  \textbf{5.02} \\
\hline
DSG w/o RBS  &   0.540 &  8.59 &  0.421 &   0.740 &  4.97 \\
DSG w/o SA &  0.522 &  8.38 &  0.421 &   0.734 &  4.83 \\
\bottomrule
\end{tabular}
  \caption{Main results.}
\label{tab:mainresult}
\end{table}

\textbf{Main Results.} The experimental results are shown in Table~\ref{tab:mainresult}. It can be observed that our DSG model outperforms all comparison models in all metrics significantly and consistently, illustrating the effectiveness and consistentness of our framework. We can draw the following conclusions. (1) The superior performance of our DSG model shows that the supportiveness scores do help alleviate the over-generation problem. (2) The DSG-A model outperforms models without any adaptor. It fails to exceed the DSG model in all metrics. It shows attention can also be used to alleviate the over-generation problem but it performs not as good as our supportiveness scores. (3) The DSG model outperforms the DSG-H model illustrating that the soft adaptor is better than the hard adaptor. (4) The ablation experiments show that both the RBS and the SA components contribute to alleviating the over-generation problem.

Specifically, the S2S model performs worse than other models. It shows that the Transformer based model works better in this task. This observation is consistent with the results observed in a lot of other similar tasks.
The S2ST model performs worse than all other transformer-based models. The reason is that it suffers severely from the over-generation problem and is very likely to generate superfluous content in the generation procedure.
The DSG-A model outperforms other models without any adaptor. This is because attention can also be regarded as a kind of supportiveness and it can be undoubtedly used to detect the over-generated words. However, since the purpose of the attention is to give weights to each input word, it is forced to assign weights to input words even no input data support the target word. As a result, it performs worse than our DSG model.
Our DSG model with a soft adaptor outperforms the DSG-H model equipped with a hard adaptor. The reason is that when the hard adaptor is used, some words are directly ignored possibly resulting in generating an incoherent target sentence. Therefore, though it outperforms other models without any supportiveness adaption, it fails to exceed our proposed DSG model.
The ablation experiment results show that both the RBS and SA components contribute to alleviating the over-generation problem. SA mainly focuses on alleviating the problem in the training phase while RBS focuses on solving it in the generation phase. They are all essential components of our model.

\begin{figure}[t]
  \centering
  \includegraphics[width=0.99\columnwidth]{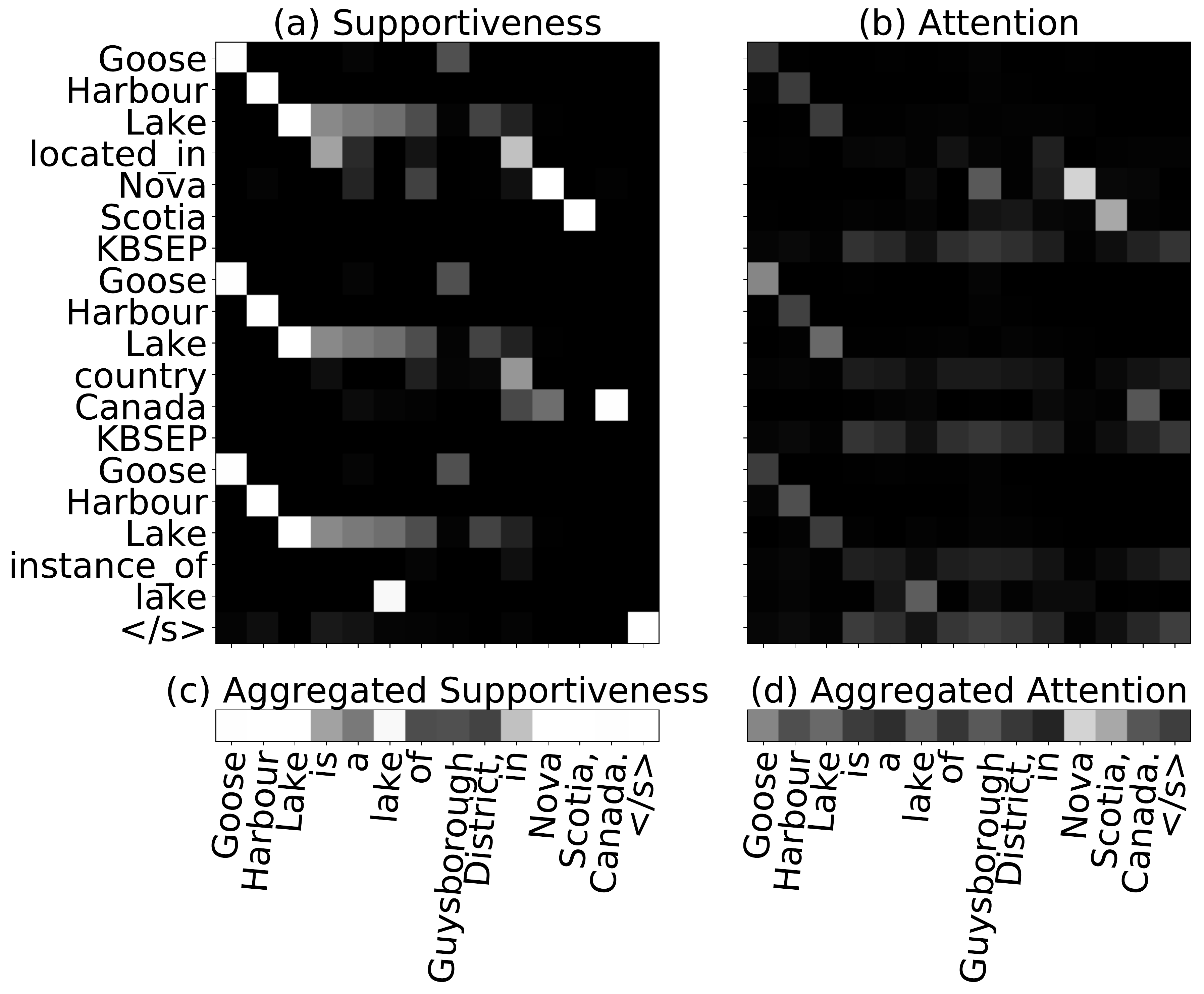}
  \caption{Comparision for supportiveness and attention. x-axis is the target text while y-axis is the given input KB triples. White stands for high score while black stands for low score. (a) and (b) show how each word in KB triples and text is aligned. (c) and (d) show the aggregated supportiveness and attention for each word in the target text.}
  \label{fig:endorse} 
  \end{figure}

  \begin{table}[t]
    \centering
    \small
  \begin{tabular}{@{~}l@{~}@{~}l@{~}@{~}l@{~}@{~}l@{~}@{~}l@{~}@{~}l@{~}}
  \toprule
  {} &    1-gram &     2-gram &     3-gram &     4-gram &     5-gram \\
  \midrule
  S2ST  &  962 &  2,313 &  3,118 &  3,425 &  3,501 \\
  DSG-A   &  894 &  2,161 &  2,934 &  3,217 &  3,290 \\
  DSG-H &  646 &  1,817 &  2,494 &  2,786 &  2,854 \\
  DSG   &  741 &  1,894 &  2,599 &  2,870 &  2,936 \\
  \bottomrule
  \end{tabular}
  \caption{N-gram statistics for over-generation error analysis.}
  \label{tab:erroranalysis}
  \end{table}

  \begin{table}[t]
    \centering
    \small
    \begin{tabular}{@{~}l@{~}@{~}|l@{~}@{~}l@{~}|@{~}l@{~}@{~}l@{~}}
  \toprule
  {} & \multicolumn{2}{@{~}c|@{~}}{S2ST} & \multicolumn{2}{@{~}c@{~}}{DSG}  \\
  \cline{2-5}
  {} & BLEU & ROUGE$_L$ &  BLEU &  ROUGE$_L$ \\
  \hline
  10k &     0.215 &        0.467 &            0.260 &               0.514 \\
  20k &     0.383 &        0.631 &            0.421 &               0.659 \\
  30k &     0.446 &        0.684 &            0.487 &               0.711 \\
  40k &     0.500 &        0.719 &            0.518 &               0.728 \\
  50k &     0.496 &        0.721 &            0.555 &               0.749 \\
  \bottomrule
  \end{tabular}
    \caption{Dataset size analysis.}
  \label{tab:small}
  \end{table}

\begin{table*}[t]
  \centering
  \small
  
\begin{tabular}{@{~}L{5.0cm}@{~}@{~}L{3.5cm}@{~}@{~}L{2.5cm}@{~}@{~}L{2.1cm}@{~}@{~}L{2.6cm}@{~}}
\toprule
{KB Triple} &  S2ST &  DSG-H & DSG &  Gold \\
\midrule
$\langle$Four Crowned Martyrs, genre, sculptural group$\rangle$,  $\langle$Nanni di Banco notable\_work, Four Crowned Martyrs$\rangle$                                                                                                                                                            &  The Four Crowned Martyrs \red{ ( also known as the Four Crowned Martyrs )} is a sculptural group \red{four} by Nanni di Banco. &                                ``Four Crowned Martyrs\blueuline{" a} sculptural \blueuline{group Nanni} di Banco. &                                 Four Crowned Martyrs is a sculptural group by Nanni di Banco. &                               Four Crowned Martyrs is a sculptural group by Nanni di Banco. \\
\hline
$\langle$Newfoundland and Labrador Route 341, located\_in, Newfoundland and Labrador$\rangle$                                                                                                                                                &                                   Route 341 is a rural road in the \red{Canadian province} of Newfoundland and Labrador. &                                       Route \blueuline{341 a} Canadian of Newfoundland and Labrador. &                                       Route 341 is a \red{settlement} in Newfoundland and Labrador. &                                       Route 341 is located in in Newfoundland and Labrador. \\

\hline
$\langle$Gaius Helen Mohiam, creator, Frank Herbert$\rangle$, $\langle$Gaius Helen Mohiam, instance\_of, fictional character$\rangle$, $\langle$Dune universe, creator, Frank Herbert$\rangle$, $\langle$Dune universe, instance\_of, fictional universe$\rangle$, $\langle$ Gaius Helen Mohiam, from\_fictional\_universe, Dune universe$\rangle$ &          Gaius Helen Mohiam is a fictional character appearing in \red{American comic books} published by Frank Herbert. &  Gaius Helen Mohiam is a fictional character \blueuline{created by Frank} \blueuline{Herberfor the Dune} \blueuline{univer}. &  Gaius Helen Mohiam is a fictional character in the Dune universe stationed by Frank Herbert. &  Gaius Helen Mohiam is a fictional character in the Dune universe created by Frank Herbert. \\
\bottomrule
\end{tabular}
  \caption{Case study. The red font stands for over-generated words while the blue underline indicates incoherent parts.}
\label{tab:casestudy}
\end{table*}

\textbf{Supportiveness Distribution Analysis.}
To give an intuitive understanding of how the supportiveness works and how it differs from the attention, we give an example comparing supportiveness distribution with the corresponding attention distribution. As shown in Fig.~\ref{fig:endorse}, it can be observed that both the supportiveness and the attention can capture the alignment relationship between words in the KB triples and words in the target sentence. The difference is that the aggregated supportiveness is high for correct words while the aggregated attention is not that significant for almost all words. The reason is that attention focuses on assigning weights to individual words with the weights summing to 1 while our proposed supportiveness score just focuses on deciding whether one word is correct or not. As a result, if one word is supported by many words in KB triples, our proposed supportiveness will be very high, while the attention may be relatively lower since its sum is fixed to 1. For example, three words ``\textit{Goose}'' in the source sequence support the same word in the target text. The supportiveness for the word is very high while the attention is low since the three words in the source sequence dissipate the attention weight.

\textbf{Over-Generation Error Analysis.} In order to perform a quantitative analysis of the over-generation problem and show how it is mitigated by our model, we conduct an over-generation error analysis in which we count all over-generated n-gram words to measure how serious the problem is. For generated sentences, we first remove all stopwords and check whether each of the remaining words appears in the given input KB triple. If it is not contained in the KB triple, we will count it as an over-generated word. The statistics are shown in Table~\ref{tab:erroranalysis}. It can be observed that the DSG-H has minimal over-generated words. This is because it directly drops all the possible over-generated words in the training. It gets the lowest over-generated words count at the cost of making the result less human-readable and thus has lower scores in other metrics like BLEU, etc. Our proposed DSG model outperforms all other models without significantly losing readability indicating that our proposed framework directly helps alleviate the over-generation problem.

\textbf{Dataset Size Analysis.} In order to explore whether our framework is capable of working on small datasets, we conduct a dataset size analysis. The results are shown in Table~\ref{tab:small}. It can be concluded that as the data size increases, all the performance of models with or without supportiveness improve noticeably. It shows that increasing data size help improve the overall scores. On the other hand, models assembled with supportiveness scores always outperform models without it. It shows that our novel architecture alleviates the over-generation problem at all scales of data size.

\begin{table}[t]
  \centering
  \small
  \begin{tabular}{l|ll}
    \toprule
    {} & Overall &  Match \\
    \hline
    S2ST  &   7.315 &  7.231 \\
    DSG-H &   7.285 &  7.331 \\
    DSG   &   7.377 &  7.569 \\
    \bottomrule
    \end{tabular}
  \caption{Human evaluation.}
\label{tab:human}
\end{table}

\textbf{Human Evaluation.} We conduct a human evaluation to eval the generation performance. We sample 130 samples from each model's generated sentences and ask human helpers to give an overall score and a match score with respect to the target sentences ranging from 1 to 10. The results are illustrated in Table \ref{tab:human}. It can be concluded from the experiment that the DSG model generates better sentences in the sense of humans.

\textbf{Case Study.} We provide a case study for several models. As shown in Table~\ref{tab:casestudy}, The S2ST model is always generating text accompanied with over-generated content while our proposed DSG model alleviates this problem significantly and consistently. When comparing the DSG-H model with the DSG model, we can find that the DSG-H model can also avoid producing over-generated content. However, it tends to remove a lot of correct words making the sentence incoherent and unreadable.
Take the last case for example: The S2ST model conveys that Gaius Helen Mohiam comes from an American comic book. However, the given input KB triple does not mention this fact. On the other hand, the DSG-H model produces ``\textit{ ... created by Frank Herberfor the Dune univer ... }'' which is even not a human-readable sentence.

\section{Related Works}
During the past few years, many tasks have been proposed to generate human-readable text from the structured data. WebNLG~\cite{gardent2017creating,gardent2017webnlg,ferreira2019neural} is proposed to describe KB triples sampled from DBPedia~\cite{auer2007dbpedia}. The E2E~\cite{novikova2017e2e,duvsek2020evaluating} task is proposed for generating restaurant reviews based on the given attributes. \citet{lebret2016neural} propose the Wikibio task to generate people's biography based on given Wikipedia infobox. \citet{fu2020open} propose to generate text based on event chains. Moreover, \citet{liang2009learning} propose to generate weather reports for weather records and \citet{wiseman2017challenges}, \citet{chen2008learning} and \citet{puduppully2019data}   propose to generate a match report according to the match briefing. All these datasets are restricted to a few domains where well-aligned data is happened to be available. No existing works are focusing on handling partially-aligned data. To solve the dataset scarcity problem, \citet{fu2020unsupervised} propose to use dual learning to train generation models based on unaligned text and knowledge triples. The model generates text based on input triples and then predict the input triples with a dual extraction model. The two models are trained alternatively with dual learning. Although \citet{DBLP:journals/corr/abs-2004-14813} proposed the ENT-DESC task aiming at generating better text description for a few entities by exploring the knowledge from KB, their focus is more on distilling the useful part from the input knowledge.

Text aligning has been studied for many years. \citet{dyer2013simple} propose the Fast Align model which is a log-linear reparameterization of IBM Model 2. \citet{legrand2016neural} propose a new score aggregation method to improve the alignment result. Moreover, attention-based models \cite{bahdanau2014neural} can also be recognized as a kind of alignment. However, these models focus on aligning source words to target words, and no existing models have been proposed to directly calculate supportiveness for generation tasks. In generation systems, \citet{fu2020dynamic} propose to dynamically align the current generation state with topics to improve the generation performance. However, it still can not directly align to the input source words.

\section{Conclusions}
In this work, we propose a new task, namely, partially-aligned Data-to-Text generation, in which we generate human-readable text based on automatically produced training data. This task is more practical and extensible to any domains. We propose a distant supervision generation framework that tackling the task. The experimental results show that our proposed model solves the over-generation problem effectively and outperforms all baseline models. Moreover, we contribute a partially-aligned dataset WITA produced by our novel automatically annotating framework which is suitable for this new task.

\section*{Acknowledgments}
The work described in this paper is substantially supported by a grant from the Research Grant Council of the Hong Kong Special Administrative Region, China (Project Codes: 14204418) and the National Natural Science Foundation of China (NSFC No. 61532010, 61732008).

\bibliography{emnlp20}
\bibliographystyle{acl_natbib}

\end{document}